# Video Key-Frame Extraction using Entropy value as Global and Local Feature

Siddu. P Algur [#1], Vivek. R [*2]

[#] *Department of Information Science Engineering, B.V. Bhoomraddi College of Engineering and Technology*
*Hubli, Karnataka, INDIA*
[1] `algursp@bvb.edu`

[*] *Department of Information Science Engineering, B.V. Bhoomraddi College of Engineering and Technology*
*Hubli, Karnataka, INDIA*
[2] `vivekr321@gmail.com`

*Abstract*— **Key frames play an important role in video annotation. It is one of the widely used methods for video abstraction as this will help us for processing a large set of video data with sufficient content representation in faster way. In this paper a novel approach for key-frame extraction using entropy value is proposed. The proposed approach classifies frames based on entropy values as global feature and selects frame from each class as representative key-frame. It also eliminates redundant frames from selected key-frames using entropy value as local feature. Evaluation of the approach on several video clips has been presented. Results show that the algorithm is successful in helping annotators automatically identify video key-frames.**

## I. INTRODUCTION

In the recent years, developments in software tools for video content management have made it possible to classify video content in efficient way and made the process of video search and retrieval even faster. Current research community concentrates on automation of video content management, overcoming the drawbacks of system involving human interaction i.e., time consumption and human perspective.

Video segmentation and key frame extraction are the bases of video analysis and content-based video retrieval. Key frame extraction, is an essential part in video analysis and management, providing a suitable video summarization for video indexing, browsing and retrieval. The use of key frames reduces the amount of data required in video indexing and provides the framework for dealing with the video content.

Video can be defined as visual representation of data. Raw video can be structured as sequence of scenes, scenes as sequence of shots and shots as sequence of frames. Most of the presented work exploits this structure of video for segmentation and key-frame extraction. Key frame is the frame which can represent the salient content and information of the shot. The key frames extracted must summarize the characteristics of the video, and the image characteristics of a video can be tracked by all the key frames in time sequence. Many methods have been developed for the selection of key frames. In a retrieval application, a video sequence is subdivided in time into a set of shorter segments each of which contains similar content. These segments are represented by representative key-frames that greatly reduce amount of data that is searched. However, key frames do not describe the motions and actions of objects within the segment. Selecting key frames of scenes allows us to capture most of the content variations, while at the same time excluding other frames which may be redundant. Choosing the first frame seems to be the natural choice, as all the rest of the frames in the scene can be considered to be logical and continuous extensions of the first frame, but it may not be the best match for all the frames in the scene.

## II. RELATED WORK

A basic rule of key frame extraction is that key frame extraction would rather be wrong than not enough. So it is necessary to discard the frames with repetitive or redundant information during the extraction [1]. Current segmentation and key-frame extraction algorithms can be classified as *temporal based segmentation* also known as *shot based segmentation* and *object based segmentation*.

### A. Shot-Based Video Segmentation

Shot-based video segmentation can be considered as a process of data abstraction, in which two major steps, i.e., temporal segmentation and key-frame extraction are usually involved. Temporal segmentation classifies one video sequence into a set of video shots using one or several frame-wise features, such as the color layout [2], entropy [3], [4] etc. Key-frame extraction implements data abstraction by selecting a set of key-frames. It is usually modeled as a typical clustering processing that divides one video shot into several clusters and selects clusters centroid as key-frames. Key-frames of each shot are extracted by using the K-means method [2]. In [5], the Gaussian mixture model (GMM) is used to model the temporal variation of color histograms in the RGB color space. Frames present in the shot can be classified into several clusters based on the feature representation. For each cluster, the frame nearest to the centroid is selected as a key-frame. The number of clusters can be determined by Bayesian Information Criterion. The main drawback of this method is that it is not able to automatically determine the number of clusters and, hence, would fail to automatically adapt the clustering to the video content.

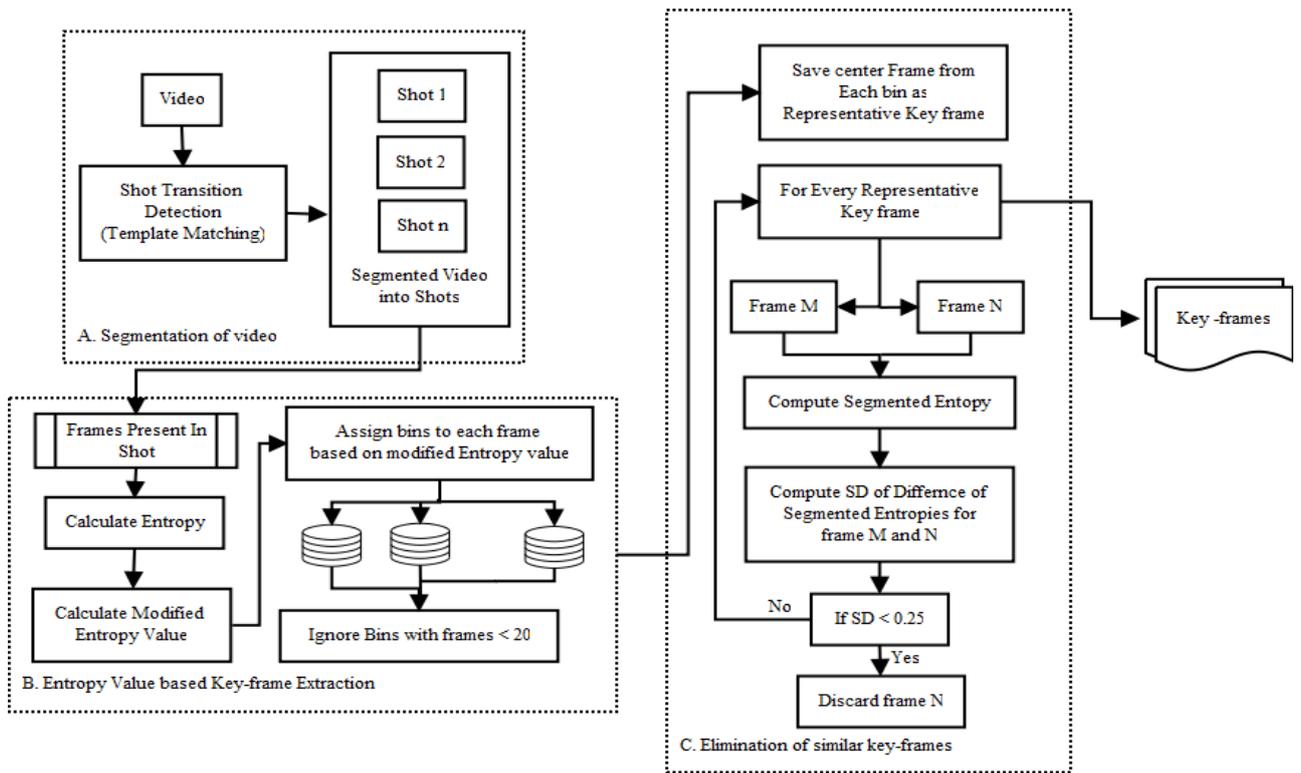

Fig. 1: System Model for Proposed Work

### B. Object-based video segmentation

Object-based video segmentation is to decompose one video shot into objects and background, which are usually application-dependent. Unlike shot-based video segmentation that has a frame as the basic unit, object-based segmentation can provide objects that represent a raw video at a higher semantic level.

Object-based video segmentation classifies a video sequence into several objects, each of which can be considered as a pattern represented by temporal or/and spatial features in a video. Current object-based video segmentation methods can be classified into three types: segmentation with spatial priority, segmentation with temporal priority, and joint spatial and temporal segmentation. More recent interests are on joint spatial and temporal video segmentation due to the nature of human vision that recognizes salient video structures jointly in spatial and temporal domains [6]. Hence, both spatial and temporal pixel-wise features are extracted to construct a multi-dimensional feature space for object segmentation. Compared with key-frame extraction methods using frame-wise features, e.g., color histogram, these approaches are usually more computationally expensive.

## III. PROPOSED WORK

The proposed approach classifies frames based on entropy values as global feature and selects frame from each class as representative key-frame. It also eliminates redundant frames from selected key-frames using entropy value as local feature. The system model for proposed work is show in Fig 1. It has three main components A. Segmentation of video into shots, B. Entropy value based key-frame extraction C. Elimination of similar key-frames present in the extracted set.

### A. Segmentation of video

Video is segmented into shots based on the detection of shot boundaries using sharp transition cut detection. The cut is defined as sharp transition between a shot and the one following. Cuts generally correspond to an abrupt change in the color and brightness pattern for two consecutive images. The principle behind this approach is that, since two consecutive frames in a shot do not change significantly in their background and object content, their overall color and brightness distribution differs little. On the other hand if we have a scene where there is a dramatically change in the color and illumination of the background it will have consequences to the color level of the image, which indicates change in the objects and background of the shot.

Segmentation of video into shots by cuts detection can be done by many methods as proposed in literature such as Histogram differences, Template matching, Edge change ratio etc. . The proposed work makes uses of template matching algorithm to segment the video. In this method two consecutive frames of the video are compared pixel by pixel and a co-relation factor between the two frames is calculated. If the co-relation factor is less than the threshold i.e., 0.9, then there is a cut present in the video, and the video is segmented into shots.

It is observed that, segmentation by template matching results in short meaning-less shots, caused by insertion of photographic effect (e.g. fades and dissolves). These shots of fades and dissolve can be merged with other shots by fixing the minimum time-frame of the shot.

*B. Entropy Value based Key-frame Extraction*

A novel algorithm for key-frame extraction based on entropy value of frames is proposed in this section. The algorithm classifies the frames in a shot into different bins.

Each bin represents a class of frames containing similar objects and background. In this algorithm entropy value is used as a global feature representing the content of the frame. Centre frame from each bin is chosen as one of the key-frame for the shot. Bins with fewer frames i.e., less than twenty frames are neglected to avoid redundant frames.

*Entropy Value:* Consider a typical frame from a video sequence, where the number of grey level is quantized to 256. Let $h_f(k)$ be the histogram of frame $f$ and $k$ the grey level, such that $0 < k < 2^b - 1$, where $b$ is the number of bits, in which the image quantization levels can be represented. If the video frame is of the class M rows N columns, then the probability of appearance of this grey level in the frame will be

$$p_f(k) = h_f(k)/(M.N) \qquad (1)$$

Entropy of the frame can be defined as sum of product of the log of the inverse probability of appearance $p_f(k)$ with probability of appearance $p_f(k)$

$$Entropy = -\sum_{k=0}^{k=\max} \log(p_f(k))/p_f(k) \qquad (2)$$

To increase the distance between the entropy bins, so that classification of each frame with varying entropy values can be performed with ease, discrete value of squared entropy values are considered (modified entropy value).

$$En_{mf} = round(En_f^2) \qquad (3)$$

Where $En_{mf}$ is modified entropy value of the frame $f$ with entropy value $En_f$. The Algorithm 1 key-frame extraction computes the modified entropy value for each frame and classifies them. New bins are initialized as when new values of entropies are computed.

*C. Elimination of similar key-frames*

It is observed that, many a times the object and background repeat in different shots of the video clip, e.g. a news reader narrating news story. This leads to one or more redundant key-frames. To eliminate these redundant key frames, a filtering step is carried out, wherein each key-frame is compared to every other key-frame to find the duplicate or near duplicate frames. To find out similarity between two frames, segmented entropy technique is applied.

In segmented entropy technique each frame is divided into 64 segments and entropy of each segment is calculated individually. In this technique, entropy is used as local feature. Through this method, variation between two frames can be monitored at segment level, which yields to more accurate identification when compared to entropy of whole frame.

To measure the dissimilarity of two frames, standard deviation of the difference of segmented entropies of two frames is computed. If standard deviation is nearing to the value Zero then the two frames are considered as similar, the second frame it eliminated as duplicate frame.

Consider two frames M and N, both frames are divided into 64 equal segments and entropies of each segments are computed. Where EnM (s1, s2…s64) corresponds to set of entropies of frame M and $E_nN$ (s1, s2…s64) corresponds to set of entropies of frame N. Entropies of each segment are obtained using "(1)".

Difference between the entropies Diff (s1, s2, s3… s64) of each segment of frame M and N are computed.

$$Diff(s_i) = En_N(s_i) - En_M(s_i) \qquad (4)$$

Standard deviation of entropy difference is computed, which signifies the dissimilarity between two frames.

$$SD = \sqrt{Variance} \qquad (5)$$

$$Variance = \sum_{i=1}^{i=64}(Diff(s_i) - mean)^2 \qquad (6)$$

### IV. EXPERIMENTAL RESULTS

The algorithm was implemented in the Open CV workspace and compared against the Entropy Difference algorithm [3]. The entropy difference algorithm has been compared with five different technique for key-frame extraction Pair-wise Pixel (P

---

ALGORITHM I: KEY-FRAME EXTRACTION

Input: video shot S which is an array of frames $S[fr_0...fr_n]$
Output: representative frames of the shot S

```
EnBins[] := 0 // Array holding modified entropy value
FrameCount[] := 0 //Array holding no. of frames present in each bins
EnFrameInBin[][] := 0 //Array holding frame number corresponding to each bin
BinCount := 0 //Number of bins
for all i such that 0 <= i <= n do
    fr := Si
    En := CalculateEntrpoy(fr)
    EnM := round(En²)  //Modified entropy value
    found := false
    for i := 0 to BinCount-1 do
        if En := EnBins[j] then
            EnFrameInBin[j][FrameCount[j]] := FrameNo(fr)
            FrameCount[j] := FrameCount[j] + 1
            found := true
        end if
    end for
    if found := false then
        EnBins[BinCount] := EnM
        EnFrameInBin[BinCount][0]] := FrameNo(fr)
        FrameCount[BinCount] := 1
        BinCount := BinCount + 1
    end if
end for
for i := 0 to BinCount-1 do
    if FrameCount[i] > 20 then
        SaveFrame(EnFrameInBin[i][FrameCount/2])
    end if
end for
```

## TABLE I
COMPARISON OF THE KEY-FRAMES EXTRACTED FROM FIVE SAMPLE VIDEO SEQUENCES USING ENTROPY VALUE

| Video | Total frames | Manually identified key frames | Entropy difference | | | Proposed Algorithm | | |
|---|---|---|---|---|---|---|---|---|
| | | | Key frame identified | Redundant frame | Missing frame | Key frame identified | Redundant frame | Missing frame |
| English News video | 3775 | 22 | 14 | 0 | 8 | 23 | 3 | 2 |
| Star trek movie clip | 2001 | 29 | 21 | 3 | 11 | 28 | 0 | 1 |
| Lord of rings trailer | 4563 | 119 | 199 | 101 | 21 | 109 | 7 | 17 |
| Lord of rings movie clip | 4002 | 67 | 45 | 0 | 22 | 85 | 24 | 6 |
| Hindi news video with graphics | 2184 | 36 | 17 | 3 | 22 | 28 | 9 | 17 |

P), x2 Test(X T), likelihood Ratio (L R), Histogram Comparison (H C) and the Consecutive Frame Difference(C f d), the details of experiment results can be found at [3]. In all the experiments reported in this section, the video streams are AVI format, with frame rate varying from 23 frames/sec to 30 frames/sec. To validate the effectiveness of the proposed algorithm, representatives' videos from News and Movie are tested. The video clips that have been selected consists of action (Lord of the rings, Star trek), conversation (news video) and inserted graphics (news videos). The video clip length varies from 1 minute sec to 4 minutes long. Initially all the video clips have been watched and key-frames from each video has picked up manually. These have been judged by a human who watches the entire video. These key-frames serve as standard against which the two different algorithms were compared and percentage accuracy was calculated regarding to how many correct key-frames each algorithm has managed to identify. Table I shows a comparison of the key-frames extracted out of the video sequences. It also allows seeing comparison between how efficiently the key-frames are extracted in terms of redundant frames and missing frames. The experimental results are shown in Fig 2. The Table II shows the deviation of the algorithm from the standard. We can see that the number of redundant frames identified by the proposed approach is comparatively lower to the entropy difference algorithm. The proposed algorithm is able to detect the presence of transient changes. On the other hand when there is inserted graphics in the video the performance of the algorithm is low, redundant frames present in the key-frame identified are comparatively higher than other video sequence.

## V. CONCLUSION

In this paper, a novel approach for automatic key-frame extraction has been presented. The proposed algorithm performance very well when the image background is distinguishable from the objects and when there is abrupt (cut) change between the shots. Algorithm works well, but with some performance loss when the video sequence contains transient changes and inserted graphics. The main advantage of the proposed algorithm is, the data loss during the key-frame extraction is minimal (number of missing frame are less) and high compactness is achieved (Number of key-frame identified/ Total frames present in the video) which is very essential property in video abstraction methods.

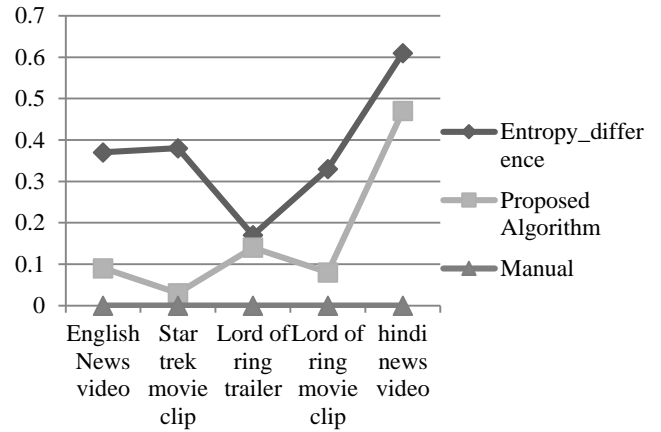

Fig. 2 Deviation from Manually identification

## TABLE II
DEVIATION FROM STANDARD (MANUALLY SELECTED) KEY-FRAMES

| Video | Entropy difference | Proposed Algorithm |
|---|---|---|
| English News video | 0.37 | 0.09 |
| Star trek movie clip | 0.38 | 0.03 |
| Lord of rings trailer | 0.17 | 0.14 |
| Lord of rings movie clip | 0.33 | 0.08 |
| Hindi news video with graphics | 0.61 | 0.47 |